\PassOptionsToPackage{table}{xcolor}
\documentclass[letterpaper]{article} 
\usepackage{aaai25}  
\usepackage{times}  
\usepackage{helvet}  
\usepackage{courier}  
\usepackage[hyphens]{url}  
\usepackage{graphicx} 
\usepackage{natbib}  
\usepackage{caption} 
\usepackage{algorithm}
\usepackage{algorithmic}

\usepackage{newfloat}
\usepackage{listings}
\DeclareCaptionStyle{ruled}{labelfont=normalfont,labelsep=colon,strut=off} 
\floatstyle{ruled}
\newfloat{listing}{tb}{lst}{}
\floatname{listing}{Listing}

\usepackage{booktabs}
\usepackage{tcolorbox}
\usepackage{xcolor}

\title{Evaluating the Meta- and Object-Level Reasoning of Large Language Models for Question Answering}
\author {
    Nick Ferguson\textsuperscript{\rm 1},
    Liane Guillou\textsuperscript{\rm 2},
    Alan Bundy\textsuperscript{\rm 1},
    Kwabena Nuamah\textsuperscript{\rm 1}
}
\affiliations {
    \textsuperscript{\rm 1}The University of Edinburgh\\
    \textsuperscript{\rm 2}Aveni\\
    nick.ferguson@ed.ac.uk, liane@aveni.ai, bundy@ed.ac.uk, k.nuamah@ed.ac.uk
}

\newcommand{\fr}{\textsc{Frank}}
\newcommand{\frl}{\textsc{Franklin}}
\newcommand{\molr}{meta- and object-level reasoning}
\newcommand{\mlr}{meta-level reasoning}
\newcommand{\olr}{object-level reasoning}

\begin{document}

\maketitle

\begin{abstract}
    Large Language Models (LLMs) excel in natural language tasks but still face challenges in Question Answering (QA) tasks requiring complex, multi-step reasoning.
    We outline the types of reasoning required in some of these tasks, and reframe them in terms of \textit{meta-level reasoning} (akin to high-level strategic reasoning or planning) and \textit{object-level reasoning} (embodied in lower-level tasks such as mathematical reasoning).
    \textsc{Franklin}, a novel dataset with requirements of \molr{}, is introduced and used along with three other datasets to evaluate four LLMs at question answering tasks requiring multiple steps of reasoning.
    Results from human annotation studies suggest LLMs demonstrate meta-level reasoning with high frequency, but struggle with object-level reasoning tasks in some of the datasets used.
    Additionally, evidence suggests that LLMs find the \olr{} required for the questions in the \frl{} dataset challenging, yet they do exhibit strong performance with respect to the \mlr{} requirements.
\end{abstract}

%

\section{Introduction}
\label{sec:intro}
Large Language Models (LLMs) have emerged as general-purpose natural-language-based task solvers.
Earlier models such as BERT \cite{devlinBERT2019} and GPT-2 \cite{radfordLanguage2019} demonstrated capability at tasks previously performed by task-specific models, such as sentiment analysis, while today's vastly increased model sizes and context windows means LLMs now find application in tasks which handle large amounts of data such as question answering (QA) over large volumes of data \cite{guuREALM2020}, and reading comprehension \cite{kociskyNarrativeQA2018}.
Alongside advances in model size, turn-based conversation has become a key mode of interaction, with commercial products like ChatGPT bringing considerable non-expert attention to the field.
QA is a primary function of these LLM-based assistants, with research efforts shifting away from simpler factoid questions and towards more complex QA varieties which require reasoning in a human-like manner (e.g., StrategyQA \cite{gevaDid2021}, CommonsenseQA \cite{talmorCommonsenseQA2019}.)
However, for a model to perform well at these QA tasks, it is not necessarily enough to simply ask a given question of a model -- supplementary techniques such as \textit{Chain-of-Thought} \cite{weiChainthought2022} are required to elicit more complex reasoning.


In this paper, we discuss the reasoning tasks expected of LLMs (section \ref{sec:reasoning-in-llms}), and some of the methods for improving LLM performance on reasoning tasks in section \ref{sec:reasoning-performance}.
We introduce our re-framing of the LLM reasoning discourse in section \ref{sec:meta-object-reasoning} in terms of the ability of LLMs to demonstrate meta- and object-level reasoning, and introduce our novel \textsc{Franklin} dataset in section \ref{sec:franklin}.
We then introduce and describe two annotation studies in section \ref{subsec:annotation-studies}, which were conducted to evaluate the reasoning capabilities of a range of state-of-the-art LLMs described in section \ref{sec:model-selection}.
These models were used to generate responses for a range of QA datasets, whose requirements we re-frame in terms of meta- and object-level reasoning in \ref{sec:dataset-selection}.
Our research questions, shown below, are addressed using results from our annotation studies in section \ref{sec:results-and-discussion}.
\begin{description}
    \item[RQ1] Do LLMs demonstrate object-level reasoning? 
    \item[RQ2] Do LLMs demonstrate meta-level reasoning? 
    \item[RQ3] Does our novel \textsc{Franklin} dataset present a challenge for LLMs? 
\end{description}

Our contributions are:
\begin{itemize}
    \item A re-framing of existing LLM reasoning discourse in terms of meta- and object-level reasoning (section \ref{sec:meta-object-reasoning}), enabling better classification of their strengths and weaknesses.
    \item The introduction of the novel \frl{} dataset, which contains meta- and object-level reasoning requirements.
    \item Evaluation of a range of state-of-the-art LLMs on datasets requiring multi-step reasoning, and discussion of the strengths and limitations of LLMs (section \ref{sec:experiment-design}).
    \item A claim that LLMs generally do not possess sufficient object-level reasoning to widely succeed at the datasets evaluated.
    \item An additional claim that LLMs are able to demonstrate meta-level reasoning consistently across the variety of datasets selected.
    \item Finally, a claim that our \frl{} dataset presents a challenge to LLMs through discussion of the low rates with which answers are provided, and the types of errors that LLMs make.
\end{itemize}

\section{Background}

We overview a selected range of reasoning tasks which LLMs are evaluated against, and techniques by which increased performance is extracted from LLMs.

\subsection{Reasoning Tasks}
\label{sec:reasoning-in-llms}

The term \textit{reasoning} is a broad cognitive concept with many forms.
Reasoning encompasses the drawing of a conclusion, using logical laws, from a set of statements (formally \cite{bundyComputer1983} or informally \cite{wasonPsychology1972}); incorporates elements of attitude revision \cite{mchughWhat2018}; and may be intuitive or explicit \cite{slomanEmpirical1996}.
Our working definition of reasoning, aiming to take into account the broad range of tasks to which the term is applied, is \textit{a task which requires some operation to infer conclusions from a set of premises}.
The range of tasks on which LLMs are evaluated represents the breadth of application of the term, with key tasks including common sense reasoning, mathematical reasoning, and symbolic reasoning.
We are particularly interested in \textit{multi-step} reasoning, which requires multiple intermediate steps of inference to draw a final conclusion.

Common sense reasoning concerns knowledge about everyday concepts which is \textit{generally accepted} by a majority of people \cite{bhargavaCommonsense2022}.
While the notion of a formal common sense logic does exist \cite{boothPTL2012}, we will discuss \textit{informal} common sense reasoning grounded in natural language.
LLMs have been shown to reflect human beliefs about generic concepts across a range of domains \cite{weirProbing2020}, and reason about physical properties of everyday objects and situations \cite{biskPIQA2020,goelHow2019}.
Similarly, LLMs encode relational data, allowing recalling of facts in a similar manner to symbolic knowledge bases (KBs) \cite{petroniLanguage2019}.
These instances reflect simple tasks where the knowledge, implicit in the parameters of an LLM, can be recalled.
However, common sense reasoning can also be a requirement for some multi-step reasoning tasks, such as the creation of a strategy for answering a question which requires multiple inferences \cite{gevaDid2021}.
As we will see, natural language-based common sense reasoning is a requirement for a variety of tasks which require multiple steps of inference to achieve a wider goal.

Mathematical reasoning concerns a model's ability to perform mathematical operations to solve problems \cite{ahnLarge2024}.
Specific tasks include arithmetic reasoning, such as addition and division, which can be expressed simply in symbolic form \cite{yuanHow2023}, or in longer-form, text-based problems \cite{cobbeTraining2021, hendrycksMeasuring2021}.
Geometry problems, which represent a conceptually harder challenge, are another example of problems requiring mathematical reasoning \cite{chenGeoQA2021}.

Symbolic reasoning tasks involve performing an action according to formal rules, albeit imitated using the prompting and output of an LLM.
This is a broader task than the mathematical reasoning task, which encompasses arithmetic and polynomial evaluation.
\citet{weiChainthought2022} describe two tasks which illustrate the challenge, although it is noted that these toy tasks are within current abilities of LLMs.
In \textit{last letter concatenation}, a model concatenates the last letters of a full name (e.g., \textit{Barack Obama} $\rightarrow$ \textit{ka}), and in \textit{coin flipping}, models are prompted to output the state of the coin after given an initial state and a number of flips.
Other examples include the emulation of formal deductive reasoning in natural language \cite{hanFOLIO2024, clarkTransformers2020}.
Some symbolic reasoning tasks are presented to the model not expressed in natural language, but symbolically.
For example, finding checkmate in a chess game -- one of the many symbolic reasoning tasks in BIG-bench \cite{srivastava2023beyond}.

There is debate over whether LLMs are \textit{actually} reasoning rather than emulating or imitating it, which itself is part of a wider debate of whether LLMs truly understand language and meaning.
Even when LLMs appear to perform reasoning tasks, it is not clear that they are reliant upon reasoning \cite{weiChainthought2022}, or if they simply using heuristics to make predictions \cite{patelAre2021}.


Many of these reasoning tasks are exemplified in QA datasets, against which LLMs are evaluated.
They also do not exist in isolation, and often require multiple steps of inference.
For example, multi-step common sense reasoning is employed to infer the steps required for solving the natural language problems in the GSM8k \cite{cobbeTraining2021} and MATH \cite{hendrycksMeasuring2021} datasets.
Similarly, basic mathematical reasoning about, for example whether a given date is before or after another date, is required for many questions in the StrategyQA dataset \cite{gevaDid2021}.

\subsection{Improving LLM Performance at Reasoning Tasks}
\label{sec:reasoning-performance}

Techniques for achieving improved performance with LLMs on reasoning tasks can broadly be placed into two categories: fine-tuning and prompt-based approaches.

Fine-tuning is a paradigm which involves taking a pre-trained model and updating the model's parameters by further training on a specific dataset \cite{devlinBERT2019}.
Fine-tuning is performative in increasing the performance of LLMs in a variety of reasoning techniques, such as arithmetic reasoning \citet{cobbeTraining2021,hendrycksMeasuring2021} and common sense QA tasks \cite{talmorCommonsenseQA2019}.

Prompting-based techniques, sometimes referred to as \textit{in-context learning}\footnote{Erroneously, as no gradient update/learning takes place.} involve taking a model's frozen weights and manipulating the content of the prompt to `externalise' the model's reasoning in the form of natural language.
\textit{Chain-of-Thought} (CoT, \citet{weiChainthought2022}) involves inserting intermediate natural language reasoning steps into the process of solving common sense, mathematical, and symbolic reasoning tasks.
A variety of closely-related techniques have since appeared and shown to increase the performance of models on reasoning tasks.
Such techniques generally involve forcing a model to generate the intermediate reasoning steps required to compute the solution to a multi-step reasoning problem.
Examples include appending \textit{``Let's think step-by-step''} to the end of a question \cite{kojimaLarge2022}; instructing models to decompose the problem into sub-problems \cite{zhouLeastmost2023}; and allowing models to explore multiple reasoning paths \cite{wangSelfconsistency2023}.

Aside from these \textit{Chain-of-Thought}-esque prompt formats, models are also prompted in an \textit{n-shot} setting in bring about improved performance.
This involves providing the model with a number of demonstrations of the task at inference time, that is, in the input to the model itself \cite{brownLanguage2020}.
For example, if instructing a model to provide English to German translations, one might include $n$ example English-German sentence pairs in the prompt itself.
$n$-shot prompting has been shown to improve the performance of models at variety of tasks \cite{radfordLanguage2019,brownLanguage2020} and indeed forms a standard component of the evaluation of LLMs on certain datasets.
For example, the performance of Llama 3.1 on GSM8k is described in an 8-shot setting, with some form of Chain-of-Though prompting applied.

\section{Reframing the Reasoning Task}

We now describe \textit{meta- and object-level reasoning} and their relevance to the task at hand, namely, multi-step QA with LLMs.
Then, building on these distinct reasoning types, we introduce a novel dataset, \frl{}, which is inspired by the \fr{} system, a QA system which employs \molr{} to infer answers to queries.

\subsection{Meta- and Object-Level Reasoning}
\label{sec:meta-object-reasoning}
Meta- and object-level reasoning are terms associated with symbolic AI, particularly the automated reasoning and proof planning domains.
We will first describe a range of definitions of these two concepts to build up a picture of their meaning.

Formally, in automated reasoning, \textit{meta-level} reasoning refers to the reasoning about the representation of a theory, while the theory itself is at the \textit{object-level} \cite{bundyComputer1983}.
\citet{bundySolving1979} uses meta-level inference to control the search of a solution to mechanics problems phrased in natural language, while object-level inference is used to compute the steps of the solution itself.
\citet{christodoulouMetareasoning1998} describes the role of meta-level reasoning as planning problem-solving strategies, controlling the use of different \textit{problem solvers} (which can be thought of as object-level reasoning components), and notes the use of \mlr{} in adapting a strategy to new knowledge which may arise during computation.
\citet{aielloReasoning1991} describe meta-level reasoning as \textit{reasoning about reasoning}, and also note its functionality in driving search strategies and the modification of a system's own behaviour.
They also, in the context of an agent-based system, distinguish the meta- and object-levels by stating that agents' world knowledge is on the object-level, while meta-level knowledge governs the links between different agents.
\citet{geneserethOverview1983} distinguishes the actions of an AI system as \textit{base-level} (or, object-level) and meta-level.
Object-level actions achieve the program's goals, while meta-level actions decide which object-level actions to perform.
\citet{nuamahALIST2023} introduce a formalism for representing knowledge in a QA system consisting of attribute-value pairs.
This formalism introduces additional attributes to the standard $\langle$\textit{subject},\textit{predicate},\textit{object}$\rangle$ triple, which may be meta- or object-level attributes.
Object-level attributes are those which encode the meaning of a factual statement, such as \textit{subject} and \textit{predicate}, while meta-level attributes capture meta-information, such as the data source for a given fact.

To summarise the above examples, meta-level reasoning approximately corresponds to the high-level planning of a solution to a problem, the decomposition of a problem into intermediate steps, and the decisions on which sub-components of a system to employ to achieve a specific task.
Reasoning on the object-level concerns the application of the sub-components.
This includes lower-level inferences, such as mathematical operations or natural language deductions, which are required to execute the intermediate steps.

We find that this delineation of reasoning tasks provides meaningful detail and structure to the discourse and classification of the reasoning tasks embodied in multi-step QA datasets on which LLMs are evaluated.
Taking GSM8k as an example, it is described as embodying a single reasoning task, namely \textit{mathematical reasoning}.
However, cursory analysis of the problems contained within the dataset show that both \molr{} are required to correctly compute answers to the questions.
In section \ref{sec:dataset-selection} we describe further examples of datasets which require \molr{}, showing that our categorisation of reasoning types as meta- or object-level generalises to a range of QA tasks, in addition to adding more fine-grained meaning.
Questions in these QA datasets are the basis for our annotation studies and evaluation of the \molr{} of the range of LLMs selected in \ref{sec:model-selection}.
However, in conducting our studies on the ability of LLMs to demonstrate \molr{}, we do not claim here that LLMs have any formal meta- or object-level reasoning component, and stress here that when applying these terms to the evaluation of LLMs, we are not evaluating a formal meta- or \olr{} component.
Rather, when we refer to LLMs as \textit{demonstrating} meta- or object-level reasoning, we refer to their ability to \textit{emulate}, or \textit{imitate} such processes via their text generation paradigm.
Our interpretation of the terms \molr{} is summarised below.
\begin{description}
    \item[Meta-level reasoning] \textit{High-level planning}. With LLMs, this is demonstrated and embodied in an informal, natural language-based decomposition of a problem in to sub-problems or intermediate steps.
    \item[Object-level reasoning] \textit{Low-level execution}. With LLMs, this is demonstrated in the execution of intermediate steps created by the \mlr{} process. Execution of these steps may require a specific task, for example, mathematical reasoning.
\end{description}

It is from this characterisation which two of our research questions, introduced in section \ref{sec:intro}, are drawn.
We revisit them here and give further detail using our above definitions.
\begin{description}
    \item[RQ1] \textit{Do LLMs demonstrate object-level reasoning?} Object-level reasoning involves low-level inferences, such as the execution of mathematical operations or natural language deduction using common sense knowledge. Can LLMs demonstrate a ability at this task across a range of datasets?
    \item[RQ2] \textit{Do LLMs demonstrate meta-level reasoning?} Meta-level reasoning governs the high-level planning and strategy for finding a solution to a problem. While we do not pretend that LLMs are employing some formal meta-level process, can LLMs demonstrate an ability to plan a solution to a range of problems as embodies in the range of datasets selected?
\end{description}

\subsection{Introducing \frl{}}
\label{sec:franklin}

\subsubsection{The \fr{} System}
To give further example of \molr{} processes in a QA setting, we will refer to the \fr{} system \cite{nuamahExplainable2020} as an example.
\fr{} is a QA system in the form of a symbolic reasoning framework which employs \molr{} in the form of a set of rules \cite{bundyUnified2022}.
These rules break queries down into sub-problems; collect data from online knowledge sources such as Wikidata, and apply mathematical operations over that data.
In \textsc{Frank}, meta-level reasoning governs the high-level approach to answering a question, including the deduction of which intermediate inferences and operations are necessary.
Object-level reasoning manifests in both the queries to knowledge bases, and in the mathematical operations applied to the data returned from knowledge bases (with the decision to use such operations taking place at the meta-level.)
Multiple lines of reasoning may be explored by the system before a final solution is assembled -- reasoning is dynamic at inference time, and not pre-determined for a given question type.

\fr{}'s functionality illustrated using questions concerning the values of geopolitical indicators belonging to different countries and regions at various points in time.
As an example, consider the question: ``\textit{Which country in Africa had the lowest population in 2012?}''.
This cannot necessarily be answered as a factoid style question by retrieving a value from a knowledge base because multiple steps of inference are required, in contrast to a question like ``\textit{What is the capital of Ghana?}'', which is simply a single fact that can be looked up.
One solution \textsc{Frank} may explore is to split \textit{Africa} into constituent countries, search for their populations in 2012, and compare values to find an answer.

The symbolic nature of this system does lead to limitations, generally as a result of the levels of hand-engineering required.
Although solutions are not hard-coded, rules which decompose queries, perform information retrieval, and aggregate data do require hand-engineering.
This lends part of the overall motivation to the project: exploring the capability of LLMs at functionality that can be performed by explicit symbolic components.

\subsubsection{The \frl{} Dataset}

\begin{table*}[!ht]
    \centering
    \begin{tabular}{llll}
        \toprule
        \textbf{Field}                   & \textbf{Description}               & \textbf{Number available} & \textbf{Example(s)}                \\
        \midrule
        \texttt{<property>}              & Geopolitical indicator             & 8                         & \textit{Female population}         \\
        \texttt{<subject>}               & Country (using ISO 3166 standard). & 249                       & \textit{Ghana}, \textit{France}    \\
        \texttt{<region>}                & ISO 3166 'sub-region'              & 16                        & \textit{Western Europe}            \\
        \texttt{<\{future,past\}\_year>} & Year in range [2008, 2030].        & 32                        & \textit{2009}, \textit{2027}       \\
        \texttt{<operator>}              & Comparison operation               & 2                         & \textit{Maximum}, \textit{minimum} \\
        \bottomrule
    \end{tabular}
    \caption{Explanation of slots which can be instantiated in \textsc{Franklin} question templates.}
    \label{tab:franklin-slots}
\end{table*}

Given this task of \molr{}, we introduce a novel dataset inspired by the \textsc{Frank} system and its exemplar domain of geopolitical indicators.
This dataset, which we call \textsc{Franklin} (\textsc{Frank} Library of Ideal Narratives) \footnote{Available in a proof-of-concept alpha version at the anonymised Github link \url{https://anonymous.4open.science/r/aaai2024-llm4plan-anon-repo-link/}.}, consists of questions, paired with template-based, natural language, step-by-step descriptions modelled on how \textsc{Frank} would nominally decompose a problem using formal deductive reasoning.
Four question templates make up the dataset, shown in figure \ref{fig:franklin-types}.

\begin{figure}[ht]
    \centering
    \begin{tcolorbox}
        \small
        \begin{description}
            \item[A. Future prediction] What will be the \texttt{<property>} of \texttt{<subject>} in \texttt{<future\_year>}?
            \item[B. Region comparison] Which country in \texttt{<region>} had the \texttt{<operator>} \texttt{<property>} in \texttt{<past\_year>}?
            \item[C. Past comparison \& future prediction] In \texttt{<future\_year>}, what will be the \texttt{<property>} of the country in \texttt{<region>} which had the \texttt{<operator>} \texttt{<property>} in \texttt{<past\_year>}?
            \item[D. Future prediction \& comparison] Will \texttt{<subject\_A>} or \texttt{<subject\_B>} have a \texttt{<operator>} \texttt{<property>} in \texttt{<future\_year>}?
        \end{description}
    \end{tcolorbox}
    \caption{The four question types which make up the \frl{} dataset.}
    \label{fig:franklin-types}
\end{figure}

Values which slots may take are detailed in table \ref{tab:franklin-slots}, and the resulting number of possible instantiations are given in table \ref{tab:franklin-numbers}.
An instantiated example of type B: region comparison is shown in figure \ref{fig:franklin-example}.
Our initial proof-of-concept release contains 400 examples, with 100 examples for each question type.

\begin{table}[ht]
    \centering
    \begin{tabular}{ll}
        \toprule
        \textbf{Question type}        & \textbf{Possible instances}       \\
        \midrule
        A. Future pred.               & 3.19$\times$10\textsuperscript{4} \\
        B. Region comp.               & 4.10$\times$10\textsuperscript{3} \\
        C. Past comp. \& future pred. & 5.24$\times$10\textsuperscript{5} \\
        D. Region comp.               & 1.59$\times$10\textsuperscript{7} \\
        \bottomrule
    \end{tabular}
    \caption{Number of possible instantiations for each question type in \frl{}.}
    \label{tab:franklin-numbers}
\end{table}

\begin{figure}[ht]
    \centering
    \begin{tcolorbox}
        \textbf{\textit{Which country in Eastern Europe had the highest energy consumption in 2019?}}
        \begin{enumerate}
            \item A list of countries located in Eastern Europe was needed. \textbf{\textit{Meta.}}
            \item 10 countries were found in Eastern Europe, including Hungary, Romania and Slovakia. \textbf{\textit{Object.}}
            \item The energy consumption for each of these countries in 2019 was needed for a comparison. \textbf{\textit{Meta.}}
            \item Data on each country's energy consumption in 2019 was found. \textit{\textbf{Object.}}
            \item The values of energy consumption were compared to each other. \textbf{\textit{Object.}}
            \item The answer to the question is the country which had the highest value. \textbf{\textit{Meta.}}
        \end{enumerate}
    \end{tcolorbox}
    \caption{Example of the \textit{region comparison} question type from the \frl{} dataset. \textit{\textbf{Step reasoning type is indicated in bold and italics.}}}
    \label{fig:franklin-example}
\end{figure}

The natural language explanations that accompany the questions in the dataset are inspired by the functionality of \fr{}, and spell out the reasoning required of any system tackling the problem, with each step having a label of meta- or \olr{}.
The reasoning type label for each step was a product of an annotation task completed by the authors.
As discussed above, \mlr{} is required to plan out how an answer to the question can be found, comprising of the setting of sub-goals and intermediate steps, and how numeric data may be aggregated to estimate an answer.
In parallel, \olr{} is required to retrieve the data, and perform mathematical operations.
The information retrieval aspect requires recalling accurate numeric data, while mathematical operations required range from simple comparisons to multiplication with 5-7 digit numbers.
For these reasons, expect both applications of object-level reasoning to be challenging for LLMs.

It should be noted that, in its current early version, the step-by-step content paired with each question takes the form of an explanation phrased as if communicating a process \textit{after} it has been performed, rather than planning out a process to be performed.
Additionally, the steps which indicate that \olr{} has taken place do not spell out the actual operations performed, but only allude to the fact that they have been performing.
These limitations, and others, will be the subject of future work as discussed in section \ref{sec:conclusion}.

The \frl{} dataset forms the basis for our third research question, which we repeat and expand on here.
\begin{description}
    \item[RQ3] \textit{Does our novel \textsc{Franklin} dataset present a challenge for LLMs?} Meta-level reasoning is required to plan out a solution to the question in terms of the necessary intermediate inferences, while object-level reasoning requires recalling factual information to high precision, and mathematical operation on said data. Can LLMs perform these actions to a sufficiently high degree?
\end{description}

\section{Experiment Design}
\label{sec:experiment-design}

In this section, we describe the annotation studies which were conducted.
We describe the content of the studies themselves; list the datasets and models which were used to generate materials for the study; and finally our evaluation metrics which we use to evaluate our research questions in section \ref{sec:results-and-discussion}.

\subsection{Annotation Studies}
\label{subsec:annotation-studies}

We firstly describe the design of the studies themselves, including the questions which were presented to participants, and the size of the studies in terms of the number of examples annotated.

\subsubsection{Study Design}
Two online human annotation studies in which we evaluated the ability of a range of LLMs to demonstrate meta- and object-level reasoning.
Four datasets, described in \ref{sec:dataset-selection} were selected.
Responses to a random sample of the questions in these datasets were generated using four LLMs, introduced in section \ref{sec:model-selection}.
For study 1, we prompted models to generate \textit{answers} to a given question, with the intention of observing object-level ability.
For study 2, we first prompted models to generate \textit{plans} for finding an answer to a given question, and followed up with an instruction to execute the plan step-by-step.
This study was designed to observe the models' meta-level reasoning ability, and also the influence of the breaking down of a problem on their ability to produce answers.
Full details of the prompts used are given in appendix \ref{app:prompts}.
In both annotation studies (built with Qualtrics\footnote{https://www.qualtrics.com/}), we asked human participants sourced from Prolific\footnote{https://www.prolific.com/} to answer questions about models' responses on a 5-point Likert scale: \textit{strongly disagree}, \textit{somewhat disagree}, \textit{neither agree nor disagree}, \textit{somewhat agree}, and \textit{strongly agree}.
In study 1, for each question, participants are required to respond to the statements below.
\begin{enumerate}
    \item The response contains an answer to the question.
    \item The response contains a clear step-by-step plan.
    \item I would be satisfied with the response if I had asked the question.
\end{enumerate}

Similarly, the below list shows the statements presented for each example in study 2.
\begin{enumerate}
    \item The response takes a rational approach to answering the question.
    \item The response contains an answer to the question.
    \item The response contains a clear step-by-step plan of how an answer can be found.
    \item Each step in the plan is visibly performed.
\end{enumerate}

\subsubsection{Study Size}
64 examples were randomly selected from the test split of each of the four datasets, and responses were generated for each example with each of the four models.
For structured responses like step-by-step plans, models often generate Markdown formatting such as bold text and bulleted lists.
We converted this formatting to HTML, which is supported by Qualtrics, so that such formatting was visible to users.
We also cleaned responses of LaTeX maths formatting to aid legibility.
This gives a total of 4$\times$4$\times$64$=$1,024 examples.
256 participants were recruited for each study, with a pre-screening process requiring participants' first language to be English.
Each participant annotated 16 examples -- one for each model/dataset combination.
This resulted in 256$\times$16 annotations evenly distributed over our 16 model/dataset combinations giving 4 annotations per example.
A pilot study was conducted with 32 participants in which 8 annotations per example were collected, however analysis of participant agreement in terms of the Standard Error of the Mean (SEM) showed that 4 annotations per example were sufficient.

\subsection{Dataset selection}
\label{sec:dataset-selection}

We selected datasets embodying tasks which require both meta- and object-level reasoning to perform.

\textbf{GSM8k} \cite{cobbeTraining2021} contains grade school mathematics problems formulated in natural language, requiring meta-level reasoning to plan the step-by-step approach, and object-level reasoning to perform the arithmetic itself.
\textbf{StrategyQA} \cite{gevaDid2021} contains questions in which the inference requirements are said to be implicit in the question.
Meta-level reasoning is required to decompose the problem and decide which intermediate inferences are necessary, while object-level reasoning is required to make deductions about relevant facts.
\textbf{HotpotQA} \cite{yangHotpotQA2018} requires meta-level reasoning to plan the intermediate steps in answering the question, and object-level reasoning to synthesise facts into an answer.
We also include the novel \textbf{\frl{}} dataset, introduced above, which requires meta-level reasoning to decompose problems into sub-problems, and object-level reasoning to retrieve information and perform mathematical reasoning.
An example of the questions in each dataset is given in figure \ref{fig:dataset-example}.

\begin{figure}[ht]
    \centering
    \begin{tcolorbox}
        \begin{description}
            \item[\frl{}] For the country in Western Europe that had the lowest GDP in 2011, what will be its urban population in 2027?
            \item[GSM8k] Peter has twice as many socks as Jack and half times as many dishes as jack. Jack collected twice as many dishes as socks in the store. If jack collected 60 dishes, calculate the total number of socks and dishes they have together
            \item[HotpotQA] Jaroslav Navratil plays for a football club that was founded in what year?
            \item[StrategyQA] Would it be common to find a penguin in Miami?
        \end{description}
    \end{tcolorbox}
    \caption{Example questions from each of the datasets employed in the study.}
    \label{fig:dataset-example}
\end{figure}

To the best of our knowledge, according to the models' white papers \cite{abdinPhi32024, dubeyLlama2024, teamGemma2024}, none of these datasets were part of a given model's pre-training or fine-tuning data, and therefore, not used to train the models themselves.

\subsection{Model selection}
\label{sec:model-selection}

Four off-the-shelf, pre-trained models were used without fine-tuning.
Meta's Llama 3.1 8B \cite{dubeyLlama2024}, Microsoft's Phi 3.5 Mini \cite{abdinPhi32024}, and Google's Gemma 2 9B \cite{teamGemma2024} were selected as examples of popular, performative, open-source models targeted towards QA and reasoning.
OpenAI's GPT-4o-mini, a closed-source model and smaller version of the flagship GPT-4o, was used as an additional comparison.
The Meta, Microsoft and Gemma models were downloaded from Huggingface and run on a local compute cluster, while the OpenAI model was queried through the OpenAI API.
Models were run in their default configurations aside from a reduction in temperature.
Temperature is a generation parameter which takes a positive number, typically on the order of 10\textsuperscript{0}.
It is proxy for `creativeness', with higher temperatures of >1 leading to 'more creative and inspiring' outputs \cite{dubeyLlama2024}.
We use a temperature of 0.2 to generate more rational, less creative responses.

\subsection{Metrics}
\label{subsec:metrics}
In assessing our claims, we refer to the metrics outlined here.

\subsubsection{Answer Failure Rate (AFR)}
\label{subsubsec:afr}

In studies 1 and 2, we asked participants to indicate whether a given response contained an answer to the question at hand (shown in questions 1 and 2 for studies 1 and 2 respectively).
The AFR is derived from these results.
It shows, for a given model/dataset combination, the proportion of questions which contain \textbf{no attempted answer} to the question at hand.
We focus on AFR rather than a standard accuracy metric because we want to observe an upper bound for a model's object-level reasoning ability -- we are less interested in absolute performance on a dataset, more so in making a comment about whether sufficient object-level reasoning is demonstrated.

To arrive at the AFR for the responses for each model/dataset combination, we took the following approach.
For each set of four annotations for a given response, we mapped the ratings on the 5-point Likert scale to a 3-point scale representing \textit{strongly/somewhat disagree}, \textit{neither agree nor disagree}, and \textit{strongly/somewhat agree}.
We then took a majority vote of these 3-point ratings to achieve a single verdict for given response.
The proportion of non-\textit{strongly/somewhat agree} verdicts gives the AFR.

\subsubsection{Rational Approach Rate (RAR)}
\label{subsubsec:rar}
In study 2, the focus is on the ability of LLMs to generate plans and, if possible, execute them to produce an answer.
In question 1 of study 2, we asked participants to indicate whether a rational approach to answering the question was present.
RAR denotes the proportion of questions which, according to annotators, contained a rational approach to solving the problem in the response.
RAR rewards models for demonstrating a general understanding of the steps required to solve a problem, even if a step-by-step plan is not explicitly created.
Given that our response generation procedure was unconstrained, we did not want to excessively penalise models for not adhering to the `step-by-step' instruction.

As with the process for obtaining AFR above, we map our results to a 3-point Likert scale and take a majority vote.
This process yields a verdict on whether a response contains a rational approach to answering the question in the case of RAR, and whether a response contains a step-by-step plan in the case of PCR.

\subsubsection{Plan Creation Rate (PCR)}
\label{subsubsec:pcr}

With PCR, we are again in the setting of study 2 where LLMs were instructed to produce a step-by-step plan as part of their response.
In question 3 of study 2, we asked participants to indicate whether a step-by-step plan was present in the response.
Similarly to RAR, PCR denotes the proportion of questions which contained a clear step-by step plan in the response.
PCR specifically targets the formatting of a step-by-step plan -- a stricter requirement than simply producing a rational approach -- models were required to clearly format a step-by-step plan.

As with the processes for the above metrics, we mapped 5-point Likert scale response to a 3-point scale, before taking a majority vote to arrive at a single verdict for the presence of a plan in a given response.
PCR indicates, the proportion of questions which contained a clear step-by-step plan according to this process.

\section{Results and Discussion}
\label{sec:results-and-discussion}

We now bring together our findings from our annotation studies, addressing our research questions using the metrics defined in section \ref{subsec:metrics}.

\subsection{Do LLMs demonstrate object-level reasoning?}
\label{sec:results-object-level}

\begin{table*}[ht]
    \centering
    \begin{tabular}{lcc|cc|cc|cc}
        \toprule
        Dataset                                        & \multicolumn{2}{c}{\frl{}} & \multicolumn{2}{c}{GSM8k}    & \multicolumn{2}{c}{HotpotQA} & \multicolumn{2}{c}{StrategyQA}                                                                             \\
        Model                                          & S1                         & \cellcolor[HTML]{C0C0C0}S2   & S1                           & \cellcolor[HTML]{C0C0C0}S2     & S1   & \cellcolor[HTML]{C0C0C0}S2   & S1   & \cellcolor[HTML]{C0C0C0}S2   \\
        \midrule
        \texttt{google/gemma-2-9b-it}                  & 0.88                       & \cellcolor[HTML]{C0C0C0}0.33 & 0.05                         & \cellcolor[HTML]{C0C0C0}0.19   & 0.28 & \cellcolor[HTML]{C0C0C0}0.20 & 0.25 & \cellcolor[HTML]{C0C0C0}0.09 \\
        \texttt{meta-llama/Meta-Llama-3.1-8B-Instruct} & 0.80                       & \cellcolor[HTML]{C0C0C0}0.05 & 0.08                         & \cellcolor[HTML]{C0C0C0}0.05   & 0.69 & \cellcolor[HTML]{C0C0C0}0.19 & 0.33 & \cellcolor[HTML]{C0C0C0}0.12 \\
        \texttt{microsoft/Phi-3.5-mini-instruct}       & 0.75                       & \cellcolor[HTML]{C0C0C0}0.16 & 0.31                         & \cellcolor[HTML]{C0C0C0}0.02   & 0.52 & \cellcolor[HTML]{C0C0C0}0.09 & 0.23 & \cellcolor[HTML]{C0C0C0}0.05 \\
        \texttt{openai/gpt-4o-mini}                    & 0.53                       & \cellcolor[HTML]{C0C0C0}1.00 & 0.02                         & \cellcolor[HTML]{C0C0C0}0.02   & 0.05 & \cellcolor[HTML]{C0C0C0}0.66 & 0.11 & \cellcolor[HTML]{C0C0C0}0.72 \\
        \bottomrule
    \end{tabular}
    \caption{Answer Failure Rate for study 1 (S1) and study 2 (S2, in the shaded boxes). Lower is better. As described in section \ref{subsubsec:afr}, this figure is the answer to the question ``\textit{What proportion of responses contained no attempt at an answer to a given question?}''}
    \label{tab:attempted-answers}
\end{table*}

Table \ref{tab:attempted-answers} shows AFR for studies 1 and 2.
The left number denotes AFR from the study 1, where LLMs provided only an answer to the question.
The right figure (shaded) denotes AFR from study 2, where LLMs were instructed to create a plan before executing that plan to answer the question.

The figures show that answers were frequently not present for a variety of model/dataset combinations.
In the study 1 setting, models overall found \frl{} the harder of the datasets, with GPT 4o-mini performing best with an AFR of 53\%.
Gemma 2's AFR of 88\% on the \frl{} dataset was the worst of any model/dataset combination.
GSM8k, with its simple arithmetic and verbose question formats, was comparatively easy compared to other datasets, with models (except in the case of Phi 3.5) failing to provide answers for less than 10\% of responses.
HotpotQA and StrategyQA, with their text-based common sense knowledge requirements, occupied a middle-ground in terms of difficulty.

AFR is consistently lower in the setting of study 2, indicating that the generation of a plan enabled models attempt answers more frequently.
This result generally aligns with the results of Chain-of-Thought-adjacent work, in which models are found to achieve better performance when prompted to decompose problems.
The exception to this decreased AFR was GPT 4o-mini, which did create plans for the questions, when asked, but specifically declined to execute that plan.
We hypothesise that this is the result of safety `guardrails' being put in place by OpenAI.

\textbf{We conclude by claiming that there is preliminary evidence that, while instructing the model to perform meta-level reasoning before answering the question results in lower AFR, models did not sufficiently, or consistently, demonstrate high levels of object-level reasoning across the range of multi-step question answering datasets.}

\subsection{Do LLMs demonstrate meta-level reasoning?}
\label{sec:results-meta-level}

\begin{table*}[ht]
    \centering
    \begin{tabular}{lcc|cc|cc|cc}
        \toprule
        Dataset                                        & \multicolumn{2}{c}{\frl{}} & \multicolumn{2}{c}{GSM8k}    & \multicolumn{2}{c}{HotpotQA} & \multicolumn{2}{c}{StrategyQA}                                                                             \\
        Model                                          & RAR                        & \cellcolor[HTML]{C0C0C0}PCR  & RAR                          & \cellcolor[HTML]{C0C0C0}PCR    & RAR  & \cellcolor[HTML]{C0C0C0}PCR  & RAR  & \cellcolor[HTML]{C0C0C0}PCR  \\
        \midrule
        \texttt{google/gemma-2-9b-it}                  & 0.95                       & \cellcolor[HTML]{C0C0C0}0.88 & 0.88                         & \cellcolor[HTML]{C0C0C0}0.91   & 0.84 & \cellcolor[HTML]{C0C0C0}0.84 & 0.95 & \cellcolor[HTML]{C0C0C0}0.94 \\
        \texttt{meta-llama/Meta-Llama-3.1-8B-Instruct} & 0.95                       & \cellcolor[HTML]{C0C0C0}0.95 & 0.95                         & \cellcolor[HTML]{C0C0C0}1.00   & 0.92 & \cellcolor[HTML]{C0C0C0}0.91 & 0.95 & \cellcolor[HTML]{C0C0C0}0.95 \\
        \texttt{microsoft/Phi-3.5-mini-instruct}       & 0.95                       & \cellcolor[HTML]{C0C0C0}0.94 & 1.00                         & \cellcolor[HTML]{C0C0C0}0.97   & 0.97 & \cellcolor[HTML]{C0C0C0}0.92 & 0.98 & \cellcolor[HTML]{C0C0C0}0.95 \\
        \texttt{openai/gpt-4o-mini}                    & 0.83                       & \cellcolor[HTML]{C0C0C0}0.88 & 1.00                         & \cellcolor[HTML]{C0C0C0}0.98   & 0.81 & \cellcolor[HTML]{C0C0C0}0.86 & 0.81 & \cellcolor[HTML]{C0C0C0}0.80 \\
        \bottomrule
    \end{tabular}
    \caption{Rational Approach Rate from study 2 (Plan Creation Rate in shaded box). Higher is better. PCR answers the question ``\textit{What proportion of responses contained a clear step-by-step plan?}'', while RAR answers the question ``\textit{What proportion of responses outlined a rational approach to answering the question?}''.}
    \label{tab:plans-rational}
\end{table*}
To answer this question, we make use of our RAR and PCR metrics described in \ref{subsec:metrics}.
Table \ref{tab:plans-rational} shows RAR and PCR across model/dataset combinations.

Results at this meta-level reasoning task show both stronger, and more consistent levels of performance at the meta-level reasoning task, frequently over 95\% for many model/dataset combinations, even for the \frl{} dataset.
As described in section \ref{subsubsec:rar} reported RAR to provide an indication that, even if the model does not produce a step-by-step plan, it still approaches the problem in a rational manner according to our annotators.
This way, we have evidence that models possess sufficient meta-level reasoning to approach the problem in an interpretable, human-understandable manner -- and we can explore ways of imposing greater structure on this in future work.

In contrast to the results for AFR in table \ref{tab:attempted-answers}, results for RAR and PCR appear to suggest that models are very competent in generating solutions to problems which take an approach which humans rate as rational, and they are similarly capable of structuring this approach in a step by step manner.
In many cases, models were able to do this for all examples in a dataset, such as in the case of Phi 3.5 on the GSM8k dataset.
Although above we suggested that the object-level reasoning in GSM8k was easier for models due its simple arithmetic and verbose questions, and that \frl{} was a harder task, we see similar levels of very high competence at the meta-level reasoning task across the range of datasets.

Again, we point out that there is speculation about whether LLMs are \textit{actually} reasoning, rather than simply \textit{imitating} it \cite{weiChainthought2022}.
We share this scepticism of models' abilities to formally reason at the meta-level and do not claim that models possess any kind of implicit, underlying, symbolic representation which this process is being completed by.
\textbf{However, we believe that our results suggest that models are able to \textit{imitate} meta-level reasoning in their text generation paradigm.}

\subsection{Does the \textsc{Franklin} dataset present a challenge for LLMs?}

Table \ref{tab:attempted-answers} shows that LLMs clearly struggled with questions from \textsc{Franklin} without first being prompted for a plan, more so than for other datasets.
While the instruction to produce a plan before answering the question lowered AFR, we referred above to analysis of a small sample of answers which show that this lower figure does convince us that models have the necessary object-level reasoning to provide \textit{correct} answers.
Different error modes are present when an answer is attempted, including data fabrication, in which the model reported non-existent values which the model claims to have found in knowledge sources; inaccurate or low-precision data, where the model reports heavily rounded or incorrect values; and incorrect arithmetic.
Examples are illustrated in figure \ref{fig:franklin-errors}.
\begin{figure}[ht]
    \centering
    \begin{tcolorbox}
        \small
        \begin{description}
            \item[Data fabrication] ``The population of Hawaii in 2017 was 1.42 million according to the World Bank.'' \textit{When this figure was manually fact-checked, no data in fact existed on the World Bank for this particular value.}
            \item[Data inaccuracy] ``The population of Togo in 2020 was 8.43 million according to the World Bank.'' \textit{When this figure was manually fact-checked, it was found to be 8,442,580.}
            \item[Incorrect arithmetic] ``12,600,000 × (1 - 0.002) = 12,492,000'' \textit{The answer is, in fact, 12,574,800.}
        \end{description}
    \end{tcolorbox}
    \caption{Examples of errors seen in the response of LLMs. The statements in quote marks are taken from the responses of LLMs, with the fact-checks appearing in italics.}
    \label{fig:franklin-errors}
\end{figure}

However, in study 2, results in table \ref{tab:plans-rational} suggest that it is not more difficult for models to \textit{plan} responses to \textsc{Franklin} questions, with plans being created with no less frequency than for 88\% of the questions in the case of GPT 4o-mini.
\textbf{From this study, we can suggest that the object-level reasoning requirements of the \textsc{Franklin} dataset presents a harder problem for LLMs, yet the success of the models at planning responses to these questions suggests that the meta-level reasoning requirements are not overwhelmingly more difficult than those of other datasets.}

\section{Conclusion}
\label{sec:conclusion}
In this paper, we have outlined reasoning tasks on which LLMs are evaluated, which are grounded in the overall setting of multi-step question answering.
These reasoning tasks have been re-framed in terms of meta- and object-level reasoning to allow us to better characterise the strengths and limitations of LLMs at these tasks.
We also introduced the novel \textsc{Franklin} dataset, which requires meta- and object-level reasoning, and which we release to the community in a proof of concept size.
Through two annotation studies, using \frl{} and three other QA datasets requiring \molr{}, we show that LLMs lack sufficient object-level reasoning to frequently provide answers to questions requiring object-level reasoning.
However, we claim that LLMs are able to sufficiently emulate meta-level reasoning in order to produce \textit{plans} for answering such questions, even for the \textsc{Franklin} dataset.
However, the object-level reasoning requirements of the \frl{} dataset were a challenge for LLMs, as demonstrated through a range of error modes.
Based on these findings, we plan continued development of \textsc{Franklin}, and evaluation of \frl{} on LLMs both off-the-shelf and fine-tuned, as well as at larger parameter counts.
These developments will consist of a broader range of question types, along with example \molr{} reasoning steps against which LLMs can be evaluated.
This work will enable us to make stronger claims about the ability of LLMs at \molr{}.
The highlighting of the weaknesses of LLMs in this and future studies will allow more targeted development of systems which address such weaknesses.

\bibliography{references}
\appendix

\section{Prompts Used to Generate Responses}
\label{app:prompts}

In study 1, responses were generated using the following simple prompt.
\begin{description}
    \item[System prompt] Answer the following question.
    \item[User prompt] \texttt{<question>}
\end{description}

In study 2, we made use of the below conversation-based prompt except in the case of Gemma 2, which does not support this feature.
When generating responses for this study using Gemma 2, we concatenated both system prompts into one and appended the question.

\begin{description}
    \item[System prompt] Create a step-by-step plan for finding the answer to the following problem. Do not answer the question. Do not perform the actions in the plan. Your only task is to outline the steps involved in a concise and clear manner.
    \item[User prompt] \texttt{<question>}
    \item[Assistant] \texttt{<response>}
    \item[System prompt] Now perform the steps in the plan you created. Use the most precise, accurate and up-to-date information available. To save space, be concise when describing the actions. Conclude by stating the answer that you reached by following the steps you outlined.
    \item[Assistant] \texttt{<response>}
\end{description}




\end{document}